\documentclass[letterpaper, 10 pt, conference]{ieeeconf}  
\usepackage{booktabs}
\usepackage{graphicx}
\usepackage{multirow}
\usepackage{bbding}
\usepackage{color}
\usepackage{url}
\usepackage{amsmath,amssymb}
\usepackage[colorlinks,linkcolor=red,anchorcolor=blue,citecolor=green]{hyperref}
\usepackage[dvipsnames]{xcolor}
\usepackage{adjustbox}

\IEEEoverridecommandlockouts                              

\title{\LARGE \bf
FlowTrack: Point-level Flow Network for 3D Single Object Tracking
}

\author{Shuo Li, Yubo Cui, Zhiheng Li, and Zheng Fang*
\thanks{The authors are all with the Faculty of Robot Science and Engineering, Northeastern University, Shenyang, China; Corresponding author: Zheng Fang, e-mail: fangzheng@mail.neu.edu.cn}
}

\begin{document}

\maketitle
\thispagestyle{empty}
\pagestyle{empty}

\begin{abstract}
    3D single object tracking (SOT) is a crucial task in fields of mobile robotics and autonomous driving. 
    Traditional motion-based approaches achieve target tracking by estimating the relative movement of target between two consecutive frames. 
    However, they usually overlook local motion information of the target and fail to exploit historical frame information effectively. 
    To overcome the above limitations, we propose a point-level flow method with multi-frame information for 3D SOT task, called FlowTrack. 
    Specifically, by estimating the flow for each point in the target, our method could capture the local motion details of target, thereby improving the tracking performance.
    At the same time, to handle scenes with sparse points, we present a learnable target feature as the bridge to efficiently integrate target information from past frames. 
    Moreover, we design a novel Instance Flow Head to transform dense point-level flow into instance-level motion, effectively aggregating local motion information to obtain global target motion.
    Finally, our method achieves competitive performance with improvements of 5.9\% on the KITTI dataset and 2.9\% on NuScenes. 
    The code will be made publicly available soon.

\end{abstract}

\section{Introduction}
3D single object tracking, an essential task in 3D computer vision, plays a crucial role in the fields of mobile robotics and autonomous driving. Given an initial target state, 3D SOT aims to capture the state of the target in subsequent frames, which can provide real-time information of target for downstream tasks like path planning and decision-making. 

Inspired by 2D image tracking \cite{SiamFC,SiamRPN,siamcar}, the previous 3D SOT methods \cite{SC3D,P2B,stnet} usually treat object tracking as a matching problem. They utilize siamese networks to extract features from the template and the current frame, then rely on appearance matching to locate the target in the current frame.
However, in scenes with object occlusion, interference from similar objects and distant target tracking, these matching-based methods face challenges due to the sparsity and disorder of point clouds. 

\begin{figure}
    \centering
    \includegraphics[width=\linewidth]{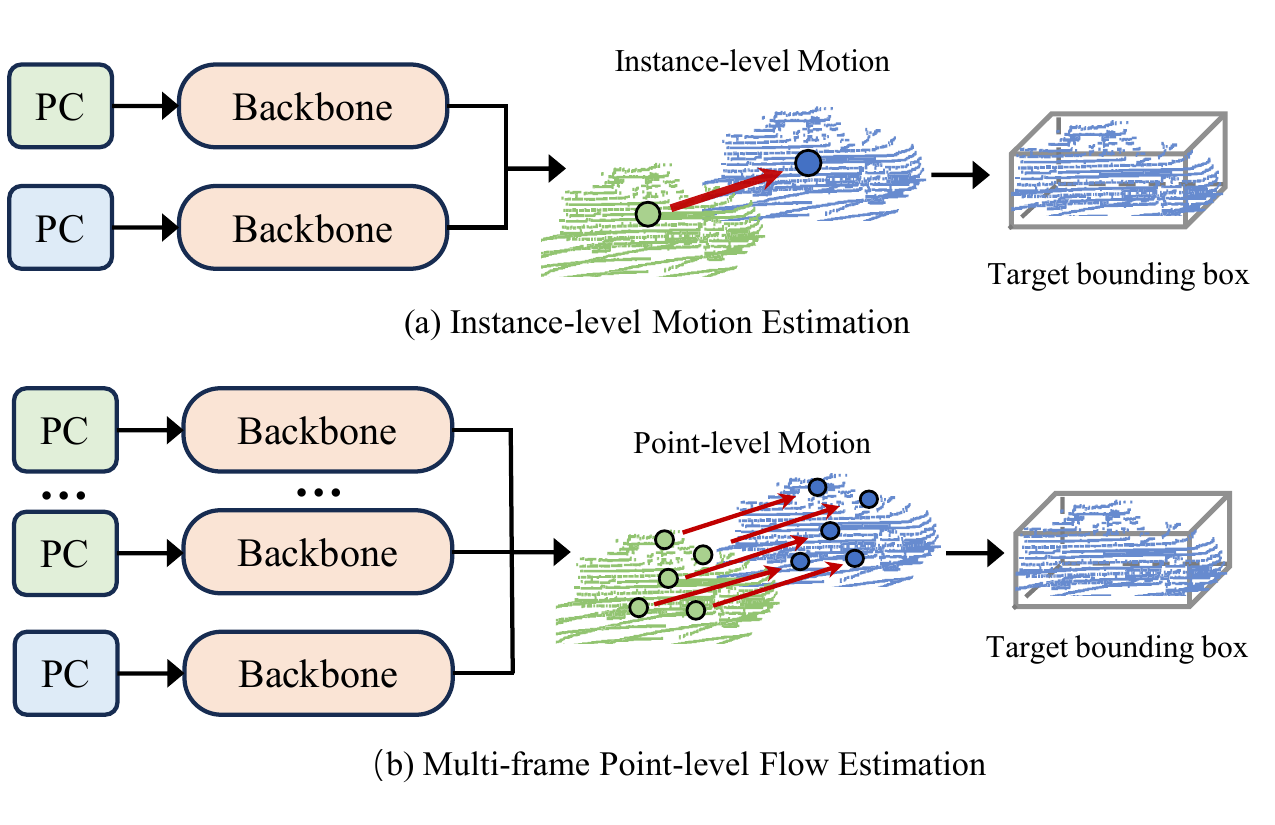}
    \vspace{-0.3in}
    \caption{
        \textbf{Comparison of the current 3D single object tracking frameworks.}
        (a) Instance-level motion estimation typically involves inputting two consecutive frames of point clouds and estimating the overall motion of the target.
        (b) Multi-frame point-level flow estimation involves inputting multiple historical frames of point clouds and estimating point-level flow within the target.
        Here, PC represents the input point clouds. 
    }
    \label{fig:1}
\end{figure}

Unlike the matching-based methods, some methods~\cite{M2,tits_1} regard the tracking task as motion prediction and locate the target by predicting the relative motion between two adjacent frames. 
This motion-based paradigm alleviates the difficulty faced by appearance matching in the above challenge scenarios. 
However, these methods only focus on instance-level target motion and ignore the local motion details of the target.  Additionally, they only utilize short-term motion cues between two frames, disregarding long-term geometry and motion information of the target from historical frames. 


To address the above problems, we introduce the idea of optical flow estimation~\cite{of4, of5} and propose a multi-frame point-level flow network, named \textbf{FlowTrack}. 
As shown in Fig.~\ref{fig:1}, unlike traditional motion-based methods, our proposed network can effectively capture the local motion details of the target by predicting the motion of each feature point in the target. 
Meanwhile, the rich geometric and motion information contained in historical frames offers valuable information supplements for estimating target motion, especially in scenes with sparse point clouds. 

Specifically, FlowTrack first extracts BEV features of consecutive frame point clouds by 3D sparse convolution~\cite{second}. Then, to extract rich motion and geometry information of the target in historical frames, we propose a Historical Information Fusion Module (\textbf{HIM}) to efficiently fuse the information of target from historical frames into the template frame. 
Subsequently, we concatenate the template and current features along the channel as flow feature and use a proposed Point-level Motion Module (\textbf{PMM}) to extract multi-scale point-level flow feature of target. 
Finally, with the aid of the Instance Flow head (\textbf{IFH}), our method accurately transform the point-level flow of the target into instance-level motion, achieving precise relative motion estimation of the target. 
Extensive experiments have demonstrated that our method achieves superior tracking performance than existing motion-based methods. 

The main contributions of this paper are as follows:
\begin{itemize}
    \item 
    We propose a multi-frame point-level flow tracking framework to achieve effective capture of the target motion information. 
\end{itemize}
\begin{itemize}
    \item 
    We design HIM, PMM, and IFH to achieve historical information fusion, target point-level flow estimation, and instance-level motion prediction respectively.
\end{itemize}
\begin{itemize}
    \item Our approach achieves competitive performance on the KITTI and nuScenes datasets. Extensive ablation studies demonstrate the effectiveness of the proposed modules.
\end{itemize}

\section{Related Work}
\label{sec:Related Work}

\subsection{3D Single Object Tracking}
Early 3D SOT methods \cite{SC3D,P2B,v2b} regard target tracking as a matching issue. They use the Siamese network to extract features of two consecutive frames and employ similarity matching for target localization. 
SC3D \cite{SC3D} is the first method to introduce the Siamese network structure into 3D SOT, enabling the capture of the target shape and subsequent generation of a series of bounding boxes within the search area. 
P2B~\cite{P2B} calculates point-level cosine similarity maps to enhance search features and adopts VoteNet \cite{votenet} for end-to-end prediction of the bounding box for target. 
PTTR~\cite{pttr} and STNet~\cite{stnet} utilize the Transformer-based structure~\cite{Transformer} to promote intra-frame information exchange and inter-frame target association. 
STTracker \cite{sttracker} goes a step further by incorporating multi-frame information to improve the performance of matching approaches. 
Although the match-based paradigm has been successful in 3D SOT, it overlooks the motion information of the target. Additionally, the sparsity and occlusion of point clouds also pose significant challenges to appearance matching.

To alleviate this problem, M$^2$-Track \cite{M2} proposes a motion-based approach, modeling the target's relative motion instead of locating target through appearance-matching in two consecutive frames. 
DMT~\cite{tits_1} designs a motion prediction module, which estimates the potential center of the target based on historical bounding boxes, and then refines it using a voting module. 
However, these methods only consider instance-level motion of target and overlook the local motion details of the target. Besides, they do not effectively utilize the rich structural and motion information in historical frame point clouds. 


\subsection{Optical Flow Estimation}
Initially, the optical flow estimation is modeled as an optimization problem, aiming to maximize the visual similarity between image pairs with the regularizations~\cite{of2,of3}. The advent of deep learning networks has significantly promoted advancements in this field. 
FlowNet \cite{flownet} is the first end-to-end convolutional neural network for optical flow estimation, but it faces challenge to handle large object displacement motions. Afterward, RAFT \cite{of4} uses loop iteration at multiple scales to achieve excellent performance with the cost of increased processing time. 
GMFlow \cite{of5} further improves the accuracy and efficiency of optical flow estimation by combining similarity matching and global attention through convolution networks. 
Furthermore, FlowFormer~\cite{flowformer} represents the pioneering method that deeply integrates transformers with cost volumes for optical flow estimation. 
Inspired by the above methods, we integrate the concept of optical flow into the 3D SOT. By estimating point-level flow of the target, our network captures detailed local motion information of the target, thus achieving more accurate and robust tracking. 

\begin{figure*}
    \centering
    \includegraphics[width=1\linewidth]{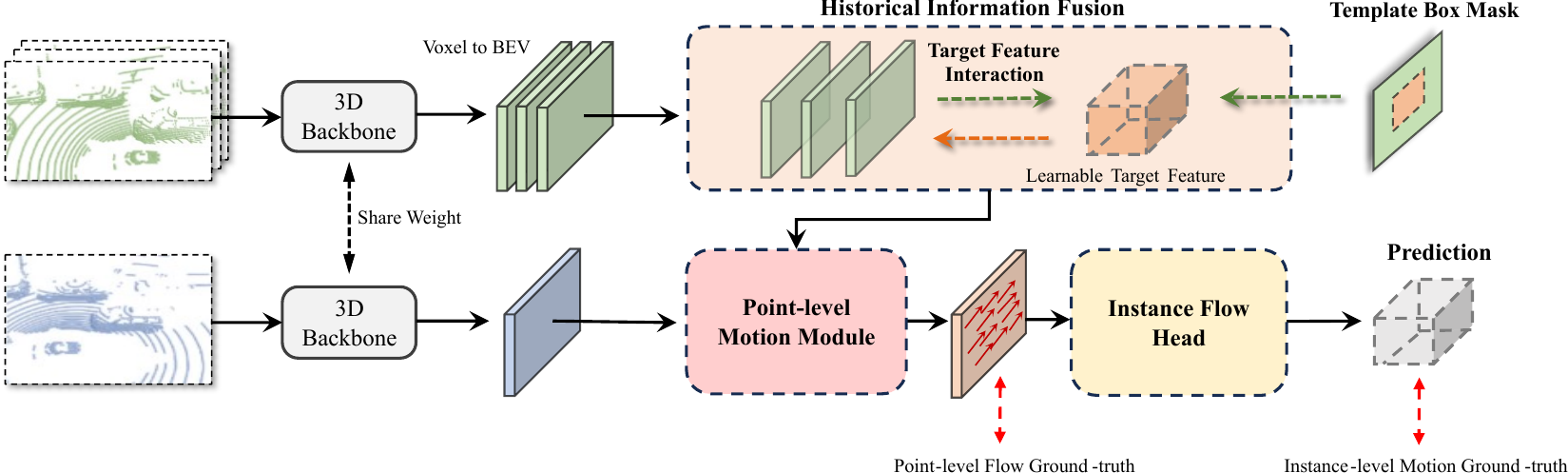}
    \caption{
    \textbf{The overall framework of FlowTrack.}
    The network comprises a 3D Feature Extraction Backbone, Historical Information Fusion Module, Point-level Motion Module and Instance Flow Head. In each frame, the point cloud is initially processed by the 3D Feature Extraction Backbone to extract voxel features and transform them into BEV. Subsequently, the Historical Information Fusion Module effectively integrates information of the target from historical frames into the template frame. Then, the Point-level Motion Module is used to obtain multi-scale point-level flow features of the target. Finally, the Instance Flow Head adaptively transforms point-level flow into instance-level target motion to obtain the final results for target tracking.
            }
    \label{fig:overall}
    \vspace{-0.1in}
\end{figure*}

\section{METHODOLOGY}
\label{sec:method}

\subsection{Framework Overview}
Given a initial bounding box $\mathcal{B}_0$, the 3D SOT task aims to estimate the $\mathcal{B}_t = (x,y,z,w,h,l,\theta)\in \mathbb{R}^{7}$ in the subsequent frame. Here, $(x, y, z)$, $(w, l, h)$, and $\theta$ demote the 3D center, size and orientation, respectively. Moreover, the size of target is assumed to remain unchanged~\cite{P2B}, thus we solely require to estimate the target center $(x, y, z)$ and orientation $\theta$. 
For motion-based paradigm, given two consecutive frames \( \mathcal{P}_t \) and \( \mathcal{P}_{t-1} \) (termed as current and template frames) and target 3D bounding box \( \mathcal{B}_{t-1} \) in \(t-1\) frame, the tracker is required to predict the relative instance-level movement between the two adjacent frames, \textit{i.e.} the translation offsets (\(\Delta x, \Delta y, \Delta z\)) and rotation angle $\Delta \theta$. Then, a rigid body transformation is applied to obtain the target bounding box $\mathcal{B}_t$ in the current frame. This process could be expressed as follows:
\begin{equation} 
    \mathcal{F}(\mathcal{B}_{t-1}, \mathcal{P}_{t-1},\mathcal{P}_t ) \mapsto
    (\Delta x, \Delta y, \Delta z, \Delta \theta)
    \label{equ:motion paradigm}
\end{equation}
\begin{equation}
\mathcal{B}_t=\operatorname{Transform}\left(\mathcal{B}_{t-1}, (\Delta x, \Delta y, \Delta z, \Delta \theta) \right)
\end{equation}

Unlike the traditional motion-based approach, we replace the coarse relative instance-level motion by estimating the fine-grained \textbf{relative point-level flow} for the target, while incorporating historical frame information. Subsequently, through a proposed Instance Flow Head, we efficiently convert point-level flow into relative instance-level motion and derive $\mathcal{B}_t$ for the current frame utilizing a rigid body transformation.
Thus, our point-level flow paradigm could be formulated as follows:
\begin{equation}
    \mathcal{F}(\mathcal{B}_{t-1},  \{ \mathcal{P}_j \}_{j=t-N}^{t})
    \mapsto \mathcal{M}
    \label{equ:flow paradigm}
\end{equation}
\begin{equation}
    \mathcal{B}_t=\operatorname{Transform}\left(\mathcal{B}_{t-1}, \text{IFH}(\mathcal{M}) \right)
    \label{equ:4}
\end{equation}
where $\mathcal{M}$ is the relative point-level from $t-1$ frame to the current $t$ frame.

Based on Eq.~\ref{equ:flow paradigm} and Eq.~\ref{equ:4}, we propose a multi-frame point-level flow network, which performs SOT by estimating the relative point-level flow of the target.
As displayed in Fig.~\ref{fig:overall}, given the point clouds from historical and current frames, we first extract BEV features for each frame. 
Afterward, we design a Historical Information Fusion Module to efficiently integrate target features from historical features, which supplements the geometric and motion information of the target. Then, we concatenate the template and current features as flow feature and feed it into the Point-level Motion Module to capture local motion details of the target at multiple scales. 
Finally, the proposed Instance Flow Head predicts the point-level flow of the target and converts it into instance-level relative motion of the target, achieving target tracking through rigid body transformation.

\subsection{Historical Information Fusion}
Traditional motion-based algorithm \cite{M2} only utilizes two consecutive frames, overlooking the rich spatial and motion information of target in history frames. 
To extract target information in historical frames, we introduce a Historical Information Fusion Module based on a learnable target feature. 
It uses the learnable target feature as a bridge and employs Transformer~\cite{trackformer} to achieve inter-frame target feature interaction, while reducing interference from background information in historical frames.

Following previous methods \cite{voxelnet}, we initially extract voxel features by voxelizing each frame's point cloud and employing three layers of 3D sparse convolution. Then, we compress the height dimension of the voxel features from the final layer to generate multi-frame BEV features $\{ \mathcal{V}_k \}_{k=t-N}^{t}$.
Additionally, to highlight the target feature in the template feature, we define a template box mask $\mathcal{Z}_{t-1} \in \mathbb{R}^{H \times W}$, where the mask value is set to 1 for grid points whose center lies within the box and 0 for others, as expressed in Eq.~\ref{equ:{Z}_{t-1}}. 

\begin{figure}
    \centering
    \includegraphics[width=1\linewidth]{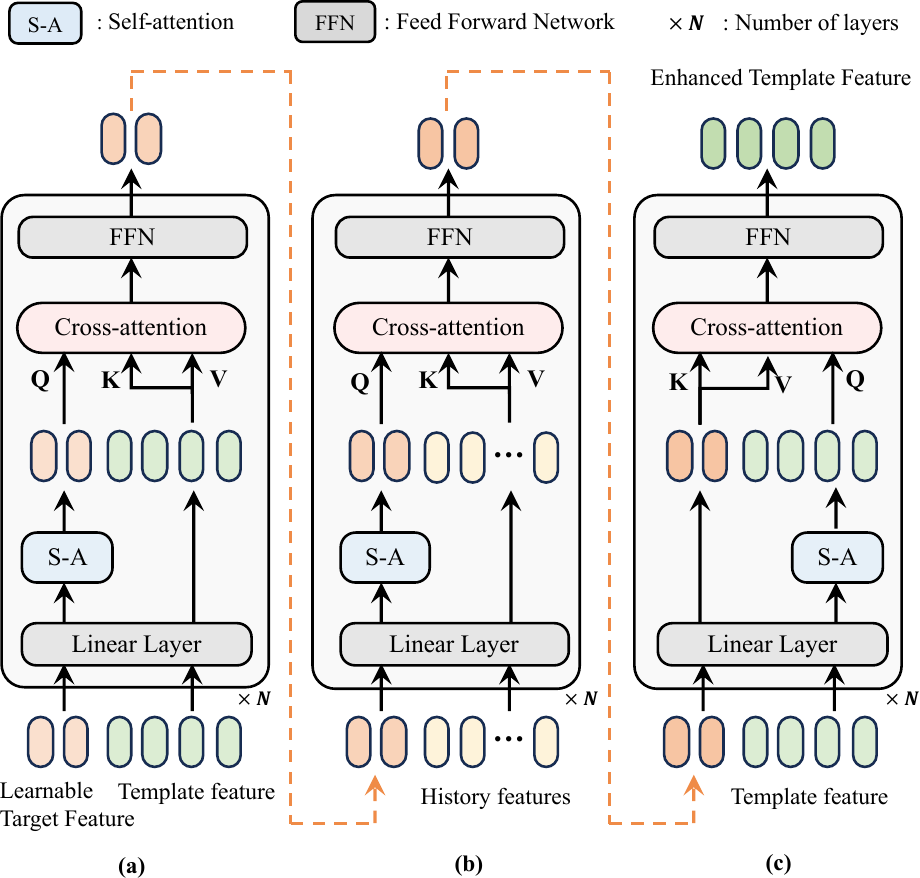}
    \caption{
    \textbf{Details of Historical Information Fusion.}
    We use learnable target feature as a bridge to facilitate information interaction between historical frames and ultimately supplement the target feature in the template frame.
    }
    \label{fig:hsitory1}
    \vspace{-0.15in}
\end{figure}

\noindent
\textbf{Learnable Target Feature Initialization}.
Based on the number of feature points occupied by the target in the template feature $\mathcal{V}_{t-1}$, 
we initially define a learnable target feature \(\mathcal{X} \in \mathbb{R}^{N \times C}\). Here, $N$ and $C$ denote the number of feature points in the target and the channel. As shown in Fig.~\ref{fig:hsitory1}(a), we update it with the target feature in the template frame by cross-attention. Specifically, we first concatenate  $\mathcal{Z}_{t-1}$ with \(\mathcal{V}_{t-1}\) along the feature channels and pass it through a 2D convolution layer to obtain mask-enhanced \({\mathcal{V}_{t-1}}'\). Then, we use cross-attention to fuse the target information from \({\mathcal{V}_{t-1}}'\) into \(\mathcal{X}\), achieving the update of \(\mathcal{X}\). 
By using \(\mathcal{Z}_{t-1} \) as guidance, we ensure \(\mathcal{X}\) learns target information rather than background information. This process can be expressed by the following formula:
\begin{equation}
    \mathcal{Z}_{t-1} = \begin{cases} 1, & \text{if point in the box} 
     \\0, & \text{others} 
    \end{cases}
    \label{equ:{Z}_{t-1}}
\end{equation}
\begin{equation}
    {\mathcal{V}'_{t-1}} = \text{Conv}( f (\mathcal{Z}_{t-1} ,  \mathcal{V}_{t-1}) )
    \label{equ:v_t-1'}
\end{equation}
\begin{equation}
    \mathcal{X}' = \text{CA}(\text{SA}(\text{Linear}(\mathcal{X})) ,\text{Linear}({\mathcal{V}'_{t-1}}),\text{Linear}({\mathcal{V}'_{t-1}}))
    \label{equ:x'}
\end{equation}
Here, \(\mathcal{X}'\) and $f$ denote the updated learnable target feature and concatenation operation, while CA and SA stand for Cross-Attention and Self-Attention. 

\noindent
\textbf{Target Feature Interaction}.
 As shown in Fig.~\ref{fig:hsitory1}(b),(c), by using the updated \(\mathcal{X}'\) as bridge, the model can interact and complement the information of the target feature in the history frames, eventually enhancing the target feature in the template frame. 
 Similar to Eq.~\ref{equ:x'}, we aggregate the target information from historical frames $ \{ \mathcal{V}_j \}_{j=t-N}^{t-2} $ into \(\mathcal{X}'\) and then use the enhanced feature \(\mathcal{X}''\) to enrich the template feature \(\mathcal{V}_{t-1}\). 
 Specifically, using \(\mathcal{X}'\) as the query, history features $ \{ \mathcal{V}_j \}_{j=t-N}^{t-2} $ as key and value, we get the enhanced target feature \(\mathcal{X}''\) by cross-attention, as shown in Fig.~\ref{fig:hsitory1}(b). Thus, \(\mathcal{X}''\) contains geometric and motion information of historical target. 
 Finally, using \({\mathcal{V}_{t-1}}\) as the query, \(\mathcal{X}''\) as the key and value, we integrate historical target information into the \(\mathcal{V}_{t-1}\) and get the enhanced template feature ${\mathcal{V}’_{t-1}}$, as shown in Fig.~\ref{fig:hsitory1}(c). 

\begin{figure}
    \centering
    \includegraphics[width=1\linewidth]{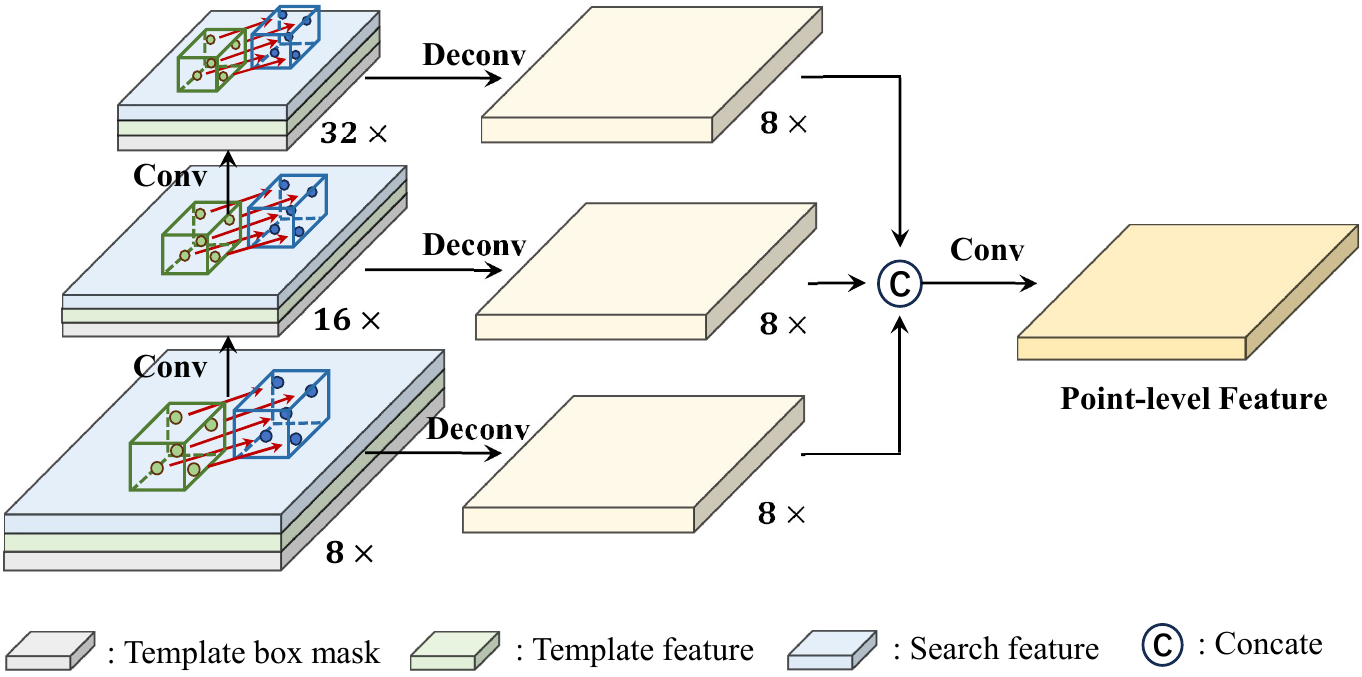}
    \caption{
    \textbf{Details of Point-level Motion Module.}
    We utilize multiple layers of convolutional layers to extract multi-scale point-level flow motion features for tracking target. These features are then concatenated along the channel dimension to obtain the final point-level flow feature.
    }
    \label{fig:PM}
    \vspace{-0.1in}
\end{figure}

\subsection{Point-level Motion Module}
Unlike motion-based methods that directly predict the instance-level motion of target, we capture local motion details within the target by predicting point-level flow, thereby achieving fine-grained motion estimation. 
However, different classes and sizes of tracked targets have different scales of motion. To fully capture their motion information, it is crucial to obtain point-level flow feature of the targets at multiple scales, as shown in Fig.~\ref{fig:PM}.

To obtain the initial point-level flow feature $F$, we first concatenate the template box mask $ \mathcal{Z}_{t-1} $, the enhanced template feature ${\mathcal{V}’_{t-1}}$ and the current feature ${\mathcal{V}_{t}}$ along the channel. 
Then, we use three layers of 2D convolution for downsampling to extract multi-scale point-level flow features $\{{F_{i}}'\}_{i=0}^{2}$ , where $i$ represents the label of the scale. Lastly, we upsample each feature in $\{{F_{i}}'\}_{i=0}^{2}$ to the size of the input feature and concatenate them along the channel to obtain final point-level flow feature ${F}''$. 
The above process can be represented by the following procedure: 
\begin{equation}
    F = f(\mathcal{Z}_{t-1}, \mathcal{V}_{t-1}, \mathcal{V}_{t}) 
    \label{equ:F}
\end{equation}
\begin{equation}
    \{{F_{i}}'\}_{i=0}^{2} = \mathcal{G}(F)
    \label{equ:F2}
\end{equation}
\begin{equation}
    {F}'' = f({ \text{Upsample}(\{{F_{i}}'\}_{i=0}^{2}}))
    \label{equ:{F}''}
\end{equation}
Here, \(\mathcal{G}\) represents a 3-layer convolution block, which performs three downsampling operations on the input \(F\). $f$ 
In Eq.~\ref{equ:F2}, \(\{{F}'\} _{i=0}^{2}\) is a collection of features from each convolutional layer output.

To train the Point-level Motion Module, we need to get point-level flow ground truth, denoted as $P_r = \{ {p_{r(i,j)}} \} \ (i \in \{1, H\}, j \in \{1, W\})$. 
Initially, we project the point cloud and target bounding box of $t$ frame to $t-1$ frame to eliminate vehicle self-motion. 
Meanwhile, we define the center point position and orientation angle of the target in these two frames as ($C_{t-1}, C_t$), ($\theta_{t-1}, \theta_t$). 
Next, to obtain the ground truth $P_r$, we need to calculate the coordinate maps $P_{t-1}$, $P_{t}$ in $t-1$ and $t$ frames, where $P = \{ {p_{i,j}} \}$ means the coordinates of each point on the feature map along the x-axis and y-axis. 
$P_{t-1}$ can be easily obtained using the target bounding box \(\mathcal{B}_{t-1}\), while $P_{t}$ is computed by transforming $P_{t-1}$ through translation and rotation, aiming to maintain the correspondence of target feature points across frames. 
Finally, we calculate $P_r$ by subtracting the value of $P_{t}$ and $P_{t-1}$. 
This process can be expressed by the following formula: 
\begin{equation}
    C_{r} = C_{t} - C_{t-1}, \theta_r = \theta_t - \theta_{t-1}
    \label{equ:C_r}
\end{equation}
\begin{equation}
    {P_t} = P_{t-1} \times {\mathcal{T}(\theta_r)} + C_{r}
    \label{equ:{R}}
\end{equation}
\begin{equation}
    {P_r} = {P_t} - {P_{t-1}}
    \label{equ:P_r}
\end{equation}
Here, $C_r$ and $\theta_r$ represent the translation and rotation of the target, respectively. 
$\mathcal{T}$ represents the transformation from orientation angle to rotation matrix. 

\begin{figure}
    \centering
    \includegraphics[width=1\linewidth]{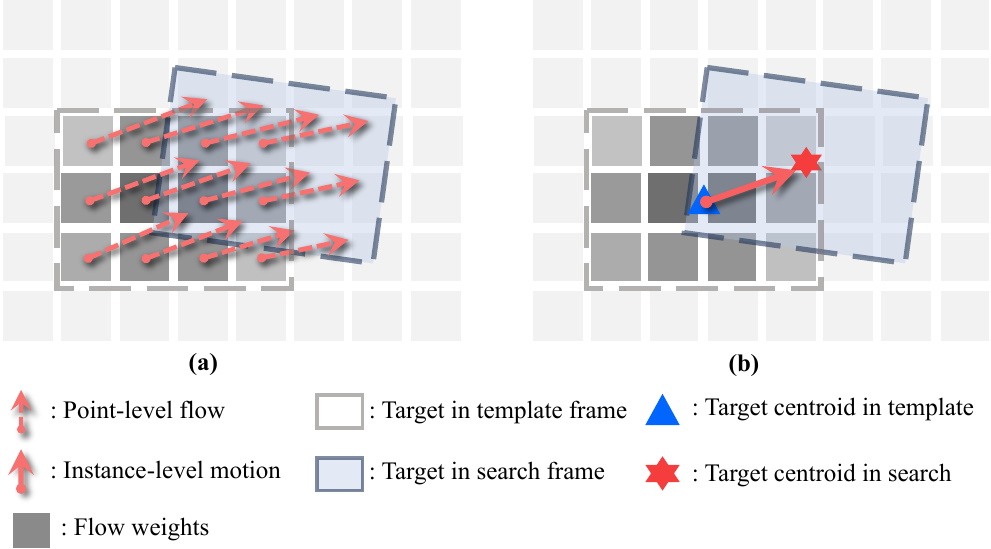}
    \caption{\textbf{The transformation from point-level flow to instance-level motion.} (a) The point-level flow map, where grey squares represent the flow weight map; (b) The instance-level motion of the target. }
    \label{flowhead}
\vspace{-0.3cm}
\end{figure}

\subsection{Instance Flow Head}
To predict the point-level flow of the target and convert it into instance-level motion, we introduce the Instance Flow Head (IFH). It predicts the point-level flow map ${M}_P \in \mathbb{R}^{H \times W \times 2}$ and the flow weight map ${W} \in \mathbb{R}^{H \times W}$. 
Here, ${M}_P$  represents the relative motion of each point of the target on the feature map, while ${W}$ denotes the weight of each point towards the overall motion. 
Specifically, using the input point-level flow feature ${F}''$, the IFH employs two separate convolutional layers to predict ${M}_P$ and ${W}$.
As shown in Fig. \ref{flowhead}, the final instance-level target motion $M_T$ is calculated by weighted averaging the value in ${M}_P$ using ${W}$. 
The formula for this process is expressed as follows:
\begin{equation}
{M_T}= \frac{\sum_{i,j} {{M}_P}_{(i,j)} \times {W}_{i,j}}{\sum_{i,j} {W}_{i,j}}
\label{equ:weight_map}
\end{equation}
Here, \(i\) and \(j\) represent coordinates on the feature map.

Through the prediction of ${W}$, the network can dynamically learn the motion weights of various points within the target, effectively distinguishing between foreground and background points. 
Then, we obtain the position coordinates \((x, y)\) of the target box along the x-axis and y-axis following  Eq.~\ref{equ:4}. For the z-axis location coordinate and the orientation \(\theta\) of the target, we directly predict them by two independent convolution layers. 

\begin{table*}[t]
\renewcommand{\arraystretch}{1.2}
\centering
\caption{{Performance comparison on the KITTI dataset.}}\label{tab:kitti}
\setlength{\tabcolsep}{4mm}{
	\begin{tabular}{c|c|c|c|cccc|c}\toprule[.05cm]
	& \multirow{2}{*}{Method} & \multirow{2}{*}{Paradigm} &\multirow{2}{*}{Source}& Car   & Ped. & Van & Cyc. &Mean \\
	& & &&\textit{6,424} &\textit{6,088} &\textit{1,248} &\textit{308} &\textit{14,068}  \\ \hline\hline\rule{0pt}{8pt}
		\multirow{13}*{\rotatebox{90}{\textit{Success / Precision}}}
		& P2B~\cite{P2B} &\multirow{10}{*}{Match} &CVPR 2020&56.2/72.8 &28.7/49.6 &40.8/48.4 &32.1/44.7 &42.4/60.0 \\
		& PTT~\cite{PTT} & &IROS 2021&67.8/81.8 &44.9/72.0 &43.6/52.5 &37.2/47.3 &55.1/74.2 \\
		& BAT~\cite{BAT} & &ICCV 2021&60.5/77.7 &42.1/70.1 &52.4/67.0 &33.7/45.4 &51.2/72.8 \\
		& PTTR~\cite{pttr} & &CVPR 2022&65.2/77.4 &50.9/81.6 &52.5/61.8 &65.1/90.5 &57.9/78.1\\
            & SMAT~\cite{SMAT} & & RAL 2022&71.9/82.4&52.1/81.5 &41.4/53.2 &61.2/87.3 &60.4/79.5 \\
            & STNet~\cite{stnet} & &ECCV 2022&\underline{72.1/84.0} &49.9/77.2 &\textbf{58.0}/\underline{70.6} &73.5/93.7 &61.3/80.1 \\
            & {GLT-T}~\cite{gltt} & & {AAAI 2023}&{68.2/82.1}& {52.4/78.8} & {52.6/62.9} & {68.9/92.1} &{60.1/79.3}\\
            & {CAT}~\cite{CAT}& & {TNNLS 2023}&{66.6/81.8} &{51.6/77.7}&{53.1/69.8} &{67.0/90.1} &{58.9/79.1} \\ 
            & STTracker~\cite{sttracker} & &RAL 2023 &{66.5/79.9} &{60.4}/\underline{89.4} &{50.5/63.6} &\underline{75.3/93.9} &62.6/82.9\\ 
            \cline{2-9}\rule{0pt}{8pt}
            & M$^2$-Track~\cite{M2}&  \multirow{3}{*}{Motion}& CVPR 2022& 65.5/80.8 &\underline{61.5}/{88.2} &53.8/\textbf{70.7}&73.2/93.5 &\underline{62.9}/\underline{83.4} \\ \rule{0pt}{8pt} 
		& DMT~\cite{tits_1} & & T-ITS 2023&66.4/79.4 &48.1/77.9 &53.3/65.6 &70.4/92.6 &55.1/75.8 \\

		& \textbf{Ours} & &- &\textbf{74.8/84.4} &\textbf{65.0/92.5} & {\underline{54.3}/65.3} &\textbf{79.1/94.7} &\textbf{68.8/86.4} \\ \cline{2-9}\rule{0pt}{8pt}
            &Improvement  &- & - &\textcolor{Green}{$\uparrow$2.7}/\textcolor{Green}{$\uparrow$0.4} &\textcolor{Green}{$\uparrow$3.5}/\textcolor{Green}{$\uparrow$3.1} &\textcolor{Red}{$\downarrow$3.7}/\textcolor{Red}{$\downarrow$5.4}  &\textcolor{Green}{$\uparrow$3.8}/\textcolor{Green}{$\uparrow$0.8} &\textcolor{Green}{$\uparrow$5.9}/\textcolor{Green}{$\uparrow$3.0}\\

            \toprule[.05cm]
	\end{tabular}}
\end{table*}
\subsection{Training Loss}
We adopt a similar approach to CenterPoint \cite{centerpoint} for predicting the target's relative motion along the x-axis and y-axis, z-axis position and orientation. Specifically, we predict the relative point-level flow map (${M}_P$), flow weight map (${W}$), z-axis location map (${M}_Z$) and orientation map (${M}_\theta $) with an 8$\times$ downsampling. Following Eq.~\ref{equ:weight_map}, the target's relative motion $M_T$ can be obtained from ${M}_P$ and ${W}$. Additionally, we assume that points within the target box have the same z-axis position and orientation. Thus, the head loss can be defined as follows:
\begin{equation}
    L_{head} = \lambda_{{M}_P}L_{{M}_P}+\lambda_{M_T}L_{M_T}+\lambda_{{M}_Z}L_{{M}_Z}+\lambda_{{M}_\theta }L_{{M}_\theta }
    \label{loss}
\end{equation}
Here, we use L1 loss for \(L_{{M}_P}\), \(L_{M_T}\), \(L_{{M}_Z}\) and \(L_{{M}_\theta}\) for attribute regression. \(\lambda_{{M}_P}\), \(\lambda_{M_T}\), $ \lambda_{{M}_Z} $ and \(\lambda_{{M}_\theta}\) are the weights for these losses respectively. To balance the network's consideration between point-level flow and flow weight map, we set the weight ratio of  \(L_{{M}_P}\) to \(L_{M_T}\) as 1:2.

\section{EXPERIMENTS}\label{sec:experiment}
\subsection{Experimental Settings}
\noindent
\textbf{Datasets \& Evaluation.}
We train and evaluate our method on KITTI dataset \cite{KITTI} and nuScenes dataset \cite{NuScenes}. KITTI dataset includes over 20,000 3D objects tracked in 21 scenes.
nuScenes dataset comprises a total of 1,000 scenes, with each frame containing approximately 300,000 points. 
We follow the One Pass Evaluation (OPE) protocol \cite{OPE1} for model evaluation, which comprises two metrics: Success and Precision. Success quantifies the degree of overlap between predicted and ground truth bounding boxes, while Precision indicates the localization accuracy of the target. 



\noindent
\textbf{Implementation Details.}
We utilize sparse 3D convolution \cite{voxelnet} widely used in 3D detection as our 3D feature extraction backbone. For our attention-based HIF Module, we follow common settings in Vision Transformer \cite{aiatrack} with the number of heads set to 8 and the number of layers set to 6. During training, we employ the Adam optimizer with an initial learning rate of 0.0003 and weight decay of 0.01 for both datasets. Training consists of 80 epochs on KITTI and 40 epochs on nuScenes, utilizing 2 NVIDIA 4090 GPUs. We set a batch size of 64 per GPU during the training phase.

\begin{table}[t]
\renewcommand\tabcolsep{3pt}
\renewcommand{\arraystretch}{1.3}
\begin{center}
\caption{Performance comparison on the NuScenes dataset. Ped and Bic is an abbreviation for pedestrian and bicycle.}~\label{tab:nuscenes}
\begin{tabular}{l|c|c|cccc|c}
\toprule[.05cm]
\multirow{2}*{}
& \multirow{2}{*}{Method} & \multirow{2}{*}{Paradigm} & Car & Ped. & Trailer & Bic. & Mean \\
& & &\textit{64,159} &\textit{33,227} &\textit{3,352} &\textit{2,292} &\textit{103,030} \\
\hline
\hline
\rule{0pt}{8pt}
\multirow{10}*{\rotatebox{90}{\textit{Success}}}
& PTT~\cite{PTT} &\multirow{6}{*}{Match} & 41.22 & 19.33 & 50.23 & 28.39 &35.52 \\ 
& P2B~\cite{P2B} & & 38.81 & 28.39 & 48.96 & 26.32 & 35.50 \\
& BAT~\cite{BAT} & & 40.73 & 28.83 & 52.59 & 27.17 & 39.98 \\
& SMAT~\cite{SMAT} & & 43.51 &{32.27} & 37.45 & 25.74 & 39.29 \\
& PTTR~\cite{PTT} & & 51.89 & 29.90 & 45.87 & - &44.44 \\ 
& STTracker~\cite{sttracker} & & \underline{56.11} & \underline{37.58} &{48.13} & 36.23 & \underline{49.43}\\
\cline{2-8}\rule{0pt}{8pt}
& M$^2$-Track~\cite{M2} & \multirow{2}{*}{Motion} &{55.85} & {32.10} & \textbf{57.61}& \underline{36.32} &{47.81} \\
& \textbf{Ours} & & \textbf{60.29} & \textbf{37.60} & \underline{55.39} & \textbf{43.74} & \textbf{52.35}  \\
\cline{2-8}\rule{0pt}{8pt}
&Improvement  &- &\textcolor{Green}{$\uparrow$4.18} &\textcolor{Green}{$\uparrow$0.02} &\textcolor{Red}{$\downarrow$2.21} &\textcolor{Green}{$\uparrow$7.42}  &\textcolor{Green}{$\uparrow$2.92}\\\hline\hline
\rule{0pt}{8pt}
\multirow{10}*{\rotatebox{90}{\textit{Precision}}}
& PTT~\cite{PTT} &\multirow{6}{*}{Match} & 45.26 & 32.03 & 48.56 & 51.19 &41.89 \\
& P2B~\cite{P2B} & & 43.18 & 52.24 & 40.05 & 47.80 & 46.10 \\
& BAT~\cite{BAT} & & 43.29 & 53.32 &44.89 & 51.37 & 46.76 \\
& SMAT~\cite{SMAT} & & 49.04 & 60.28 & 34.10 & 61.06 & 52.45 \\
& PTTR~\cite{PTT} & & 58.61 & 45.09 & 38.36 & - &53.48 \\ 
& STTracker~\cite{sttracker} & & \underline{69.07} & \textbf{68.36} & {55.40} & \textbf{71.62} & \underline{68.55}\\
\cline{2-8}\rule{0pt}{8pt}
& M$^2$-Track~\cite{M2} &\multirow{2}{*}{Motion} & {65.09} & {60.92} & \underline{58.26} & {67.50} & {66.77} \\
& \textbf{Ours} &  & \textbf{71.07} & \underline{67.64} & \textbf{62.70} & \underline{71.43} & \textbf{69.70} \\
\cline{2-8}\rule{0pt}{8pt}
&Improvement  &- &\textcolor{Green}{$\uparrow$2.00} &\textcolor{Red}{$\downarrow$0.71} &\textcolor{Green}{$\uparrow$4.44} &\textcolor{Red}{$\downarrow$0.19}  &\textcolor{Green}{$\uparrow$1.15}\\
\toprule[.05cm]
\end{tabular}
\end{center}
\vspace{-2.0em}
\end{table}

\subsection{Quantitative Results}
\noindent
\textbf{Results on KITTI.}
As shown in Tab.~\ref{tab:kitti}, our method outperforms the best matching-based paradigm, STTracker~\cite{sttracker}, by \textbf{6.2\%/3.5\%} in Success/Precision and surpasses the best motion-based paradigm, M$^2$-Track~\cite{M2}, by \textbf{5.9\%/3.0\%} in Success/Precision. 
Specifically, Pedestrian targets typically present sparser point clouds and are more susceptible to interference from similar pedestrians. 
This poses challenges for appearance-based matching methods~\cite{stnet, BAT, pttr}. M$^2$-Track~\cite{M2} partially addresses this challenge by predicting instance-level relative motion for tracking task. However, It overlooks the local motion details of the target and prior information from historical frames, resulting in suboptimal performance. 
Our method further transforms tracking task into a point-level flow estimation task, which effectively captures the local motion details of target, thereby improving the tracking performance in similar interference scenes. 
Meanwhile, in sparse tracking scenes, our method leverages the target information from multiple historical frames, resulting in a performance enhancement of \textbf{4.6\%/3.1\%} in Success/Precision for the pedestrian category.

\noindent
\textbf{Results on nuScenes.}
We validated our method on the larger tracking dataset nuScenes, and the result is shown in Tab.~\ref{tab:nuscenes}. 
Our method surpasses the best motion-based method\cite{M2} by \textbf{4.54\%/6.12\%} in Success/Precision and outperforms the best match-based method~\cite{sttracker} by \textbf{2.92\%/1.15\%} in Success/Precision. 
In particular, for the largest number of categories, vehicle, our method achieves outstanding performance improvements, outperforming the suboptimal method by \textbf{4.18\%/2.00\%} in Success/Precision.  
This is mainly attributed to the accurate capture of vehicle local motion details by the PMM module and the effective aggregation of historical information by the HIM module. 
Meanwhile, we observe that our model does not demonstrate superior performance when tracking large-size targets, such as Vans in KITTI and Trailers in nuScenes. We believe that the larger size of these targets results in significant variations in local point motion, thereby affecting the final instance-level motion estimation.


\begin{figure}[t]
    \centering
    \includegraphics[width=0.95\linewidth]{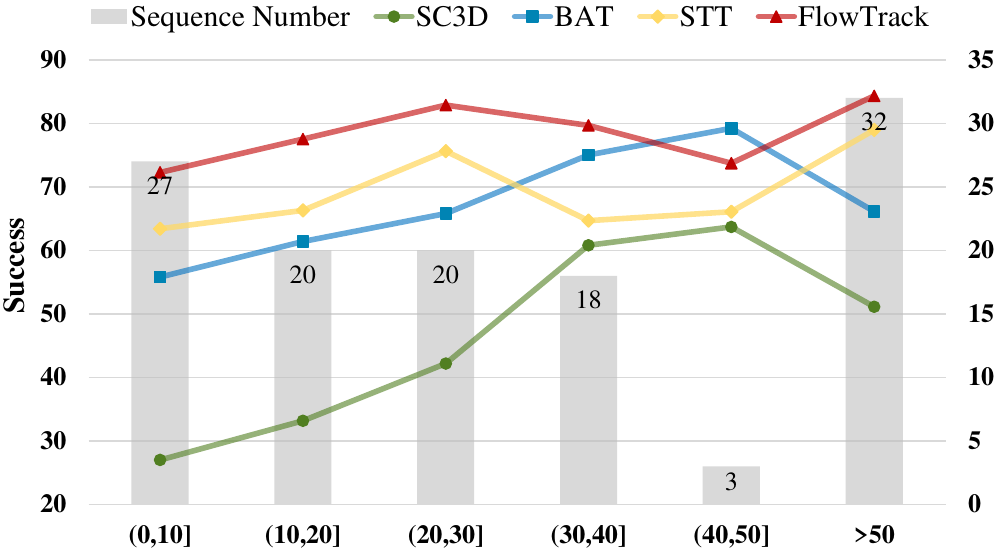}
    \caption{Effect of the number of points for the target in the initial frame. In most cases, our FlowTrack significantly outperforms other methods.}
    \label{fig:first_number}
\vspace{-0.3cm}
\end{figure}


\noindent
\textbf{Robustness to Sparseness.}
In Fig.~\ref{fig:first_number}, consistent with prior studies~\cite{SC3D,BAT,sttracker}, we display tracking performance under the different number of points for the target in the initial frame. 
our method surpasses alternative approaches in the majority of scenarios, specifically those with fewer than 40 points. This result confirms the distinct advantage of our method when dealing with sparse scenarios.

\noindent
\textbf{Running speed.}
As shown in Tab.~\ref{table:speed}, despite the introduction of multiple frames in our method, we still maintain a decent inference speed of 33.2 FPS for the model.



\begin{table}[t]
\centering
\caption{Comparison of the running speeds.}\label{table:speed}
\setlength{\tabcolsep}{1.4mm}{
\begin{tabular}{c|cccc}
\toprule[.05cm]
Method &PTT\cite{PTT} &SMAT\cite{SMAT} &STNet\cite{stnet} & M$^2$-Track\cite{M2} \\ 
FPS    & 40.0 & 17.6 &35.0 & 57.0\\ 
\hline\hline
Method &GLT-T~\cite{gltt} &CAT~\cite{CAT} &STTtracker~\cite{sttracker}& \textbf{Ours}  \\
FPS & 30.0 &45.3 &23.6 &33.2 \\
\toprule[.05cm]
\end{tabular}}
\end{table}

\begin{table}[t]
\vspace{-0.1in}
\renewcommand\tabcolsep{2.7mm}
\vspace{0.1in}
\caption{Ablation of our components.HIM, PMM and IFH respectively means Historical Information Fusion, point-level Motion Module and Instance Flow head}\label{tab:main_component}
\begin{adjustbox}{center}
\begin{tabular}{c|ccc|c|c}\toprule[.05cm]
& HIM          & PMM          & IFH         & 3D Success    & 3D Precision    \\ 
\cline{2-6}  \hline 
\midrule
A1 &\checkmark &\checkmark &\checkmark &74.8 &84.4 \\
A2 &           &\checkmark &\checkmark &71.2 &80.6 \\
A3 &\checkmark &           &\checkmark &63.6 &73.5 \\
A4&\checkmark &\checkmark  &           &71.4 &81.6 \\
\toprule[.05cm]
\end{tabular}
\end{adjustbox}
\vspace{-0.5cm}
\end{table}

\subsection{Qualitative analysis}
\begin{figure*}[t]
    \centering
    \includegraphics[width=\linewidth]{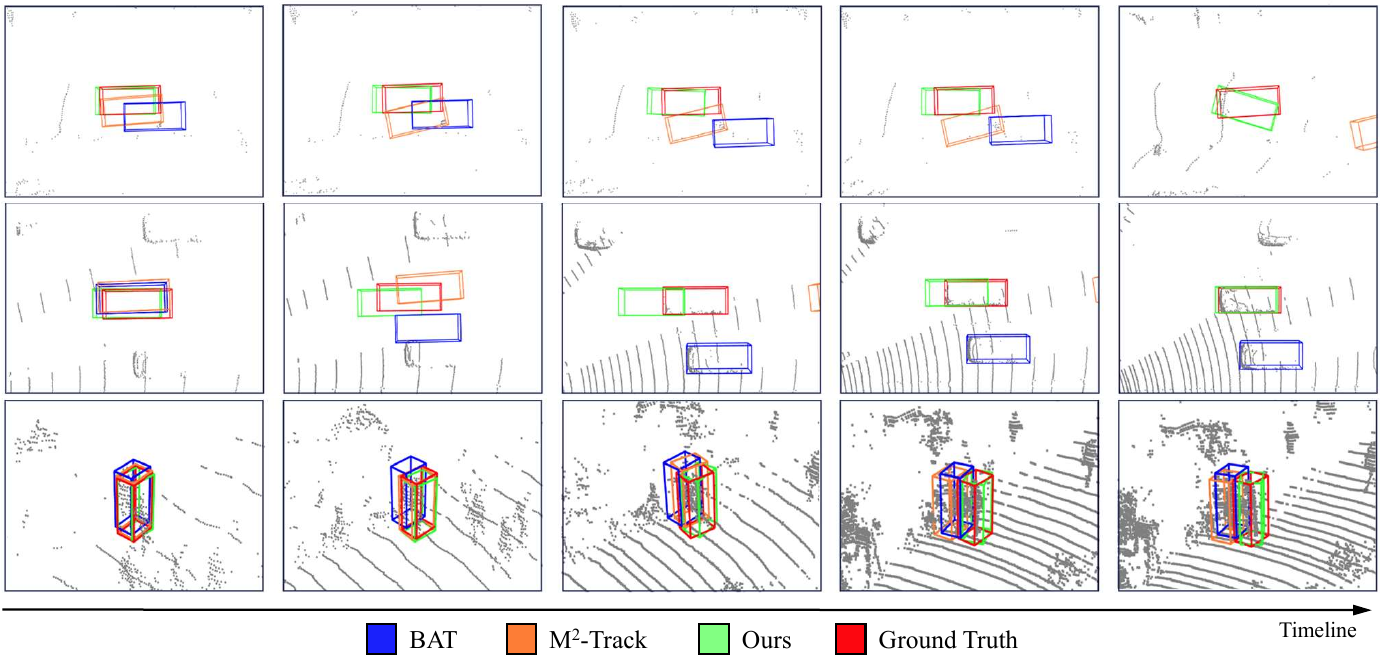}
    \caption{\textbf{Visualization of tracking results on KITTI dataset.} From top to bottom, we compare our FlowTrack with BAT, M$^2$-Track and show the cases of Car and Pedestrian categories. 
    1$^{st}$: Point sparsity cases; 2$^{st}$: Occlusion cases; 3$^{st}$: Interference from similar targets. 
    }
    \label{fig: Visulization}
    \vspace{-0.05in}
\end{figure*}

To better illustrate our method's advantages in various scenes, we visualize some challenging scenes from the KITTI dataset, as shown in Fig.~\ref{fig: Visulization}. In sparse and occluded scenes (the 1$^{st}$ and 2$^{st}$ rows), BAT, utilizing appearance matching, and M$^2$-Track, relying on instance-level motion estimation, gradually lose track of the target. 
In contrast, our method utilizes the abundant target information from historical frames to provide prior knowledge for model tracking, which aids in recapturing the target once the point clouds become denser. 
In scenes with interference from similar objects (3$^{st}$ row), our point-level flow motion estimation captures finer local motions, thereby better distinguishing interfering objects and enhancing tracking performance. 

\subsection{Ablation Study}
To validate the effectiveness of our proposed modules, including Historical Information Fusion(HIM), Point-level Motion Modole(PMM) and Instance Flow Head(IFH), we conduct ablation studies on the Car of the KITTI dataset.

\noindent
\textbf{Model Components.}
Firstly, we conduct an ablation study on the main components of our method to explore the effects of each module. As shown in Tab.~\ref{tab:main_component}, A1 represents all network models of our method. 
In A2, the removal of the HIM results in a performance decrease of 3.6\%/3.8\% in success/precision. 
This result demonstrates that the HIM effectively aggregates historical information, thereby improving the tracking performance of the network. In A3, when we remove the PMM, the model's performance shows a significant decline, with success and precision decreasing by 11.2\% and 10.9\% respectively. 
This demonstrates the critical importance of multiscale point-level flow in target motion prediction, as it comprehensively captures the local motion details of the target. 
In A4, we replace the IFH with a traditional center head~\cite{centerpoint} and observe a decrease of 3.4\%/3.3\% in Success/Precision, respectively. 
It is evident that IFH accurately transforms dense point-level flow into instance-level motion, thereby enhancing tracking performance.

\begin{figure}[t]
\centering
\includegraphics[width=0.97\linewidth]{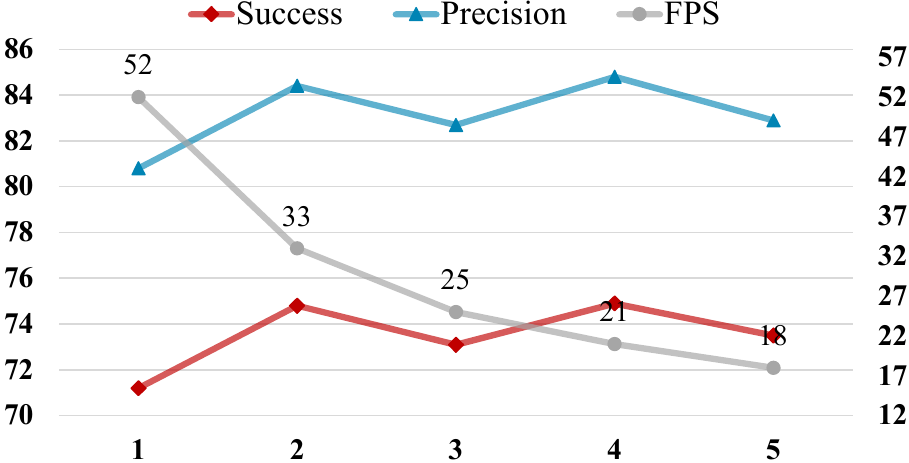}
\caption{\textbf{Performances with different number of historical frames.} We input different numbers of historical frames, from $t-1$ to $t-5$. Our method could achieve higher performance with more previous frames.}
\label{fig: input_frames}
\vspace{-0.1in}
\end{figure}

\noindent
\textbf{Input History Frames.}
To explore the impact of the number of historical frames $\{\mathcal{P}_j \}_{j=t-N}^{t-1}$ on the model's performance, we conducted an ablation experiment by adjusting the input historical frames. As illustrated in Fig.~\ref{fig: input_frames}, our method showed significant enhancement as the historical frames increased from 1 to 2, improving performance by 3.6\%/3.5\%. However, as the historical frames increased further, performance fluctuated around 73.8\%/83.5\%. 
Earlier historical frames do not always contribute positively to the tracking of the target due to their inclusion of outdated target states. 
Additionally, an excessive number of historical frames can increase computational complexity and reduce the model's speed. Hence, to balance speed and performance, we adopt the default setting of two history frames as input for our method, maintaining a performance of 33 FPS.

\noindent
\textbf{Comparison in Historical Frames Fusion.} 
As shown in Tab.~\ref{tab:history comp}, We conducted experimental comparisons on different historical frames fusion methods to validate the effectiveness of the HFM. In B1, we directly concatenate all historical frames along the feature channels. 
In B2, we employ a cross-attention mechanism, using ${\mathcal{V}_{t-1}}$ as the query and $\{\mathcal{V}_k\}_{k=t-N}^{t-2}$ as key and value. 
In B3, the HIF uses the learnable target feature as a bridge to achieve efficient information interaction between tracking targets across frames. 
The performance of the concatenation in B1 and the attention mechanism in B2 has decreased by 3.0\%/2.9\% and 2.7\%/2.7\% when compared to the HIF in B3. 
We believe that the operation of directly concatenating features from historical frames in B1 is overly simplistic for information aggregation, as it overlooks the alignment of target features across multiple frames. 
While the cross-attention in B2 achieves inter-frame information interaction, the feature interaction across the entire feature maps often introduces interference from background information. By using the learnable target feature, the HIF can more effectively focus on the feature fusion of target information within the frame, significantly reducing background information interaction and achieving superior results.

\begin{table}[t]
\renewcommand\tabcolsep{11pt}
\renewcommand{\arraystretch}{1.3}
\centering
\caption{Comparison of historical frames fusion methods}\label{tab:history comp}
\begin{tabular}{c|c|c|c}
\toprule[.05cm]
{} & {Methods} & {3D Success} & {3D Precision} \\ \hline \hline
\rule{0pt}{8pt}
B1 & Concatenation         & 71.8 & 81.5 \\
B2 & Attention       & 72.1 & 81.7 \\
B3 & HIF  & 74.8 & 84.4 \\
\toprule[.05cm]
\end{tabular}
\end{table}

\begin{table}[t]
\vspace{-0.1in}
\renewcommand\tabcolsep{9pt}
\renewcommand{\arraystretch}{1.3}
\centering
\caption{Comparison of head methods}\label{tab:head}
\begin{tabular}{c|c|c|c}
\toprule[.05cm]
{} & {Methods} & {3D Success} & {3D Precision} \\ \hline \hline
\rule{0pt}{8pt}
C1 & Center Head       & 71.4 & 81.1\\
C2 & MLP       & 70.0 & 80.4 \\
C3 & Instance Flow Head              & 74.8 & 84.4 \\
\toprule[.05cm]
\end{tabular}
\vspace{-0.5cm}
\end{table}

\noindent
\textbf{Comparison in Head Methods.} 
In the Tab.~\ref{tab:head}, we compared the impact of different prediction heads on the model's performance. In comparison to our proposed Instance Flow head, adopting the Center Head, MLP led to performance drops of 3.4\%/3.3\% and 4.8\%/4.0\%, respectively. The prediction heads in C1 and C2 do not effectively consider the transformation relationship from dense point-level flow motion to instance-level target motion, thereby leading to a decline in performance. In contrast, our IFH achieves accurate transformation between the two by adaptively predicting flow weight maps.


\section{Conclusion} 
\label{sec:conclusion}
In this paper, we propose a multi-frame point-level flow tracking network, FlowTrack, which achieves target tracking by capturing the local motion information of the target. Simultaneously, 
the proposed HIF and IFH effectively aggregate target information across temporal dimensions and convert the point-level flow of target to instance-level motion. 
Extensive experiments demonstrate that our method outperforms previous matching and motion paradigm methods, achieving competitive performance on the KITTI and NuScenes datasets.

\bibliographystyle{ieeetr}
\bibliography{root}

\end{document}